\documentclass{article}

\usepackage{PRIMEarxiv}

\usepackage[utf8]{inputenc} 
\usepackage[T1]{fontenc}    
\usepackage{hyperref}       
\usepackage{url}            
\usepackage{booktabs}       
\usepackage{amsfonts}       
\usepackage{amsmath} 
\usepackage{nicefrac}       
\usepackage{microtype}      
\usepackage{lipsum}
\usepackage{graphicx}
\usepackage{microtype}      
\usepackage{multirow}       
\graphicspath{{media/}}     

\title{Hierarchical MLANet: Multi-Level Attention for 3D Face Reconstruction from Single Images
}

\author{
  Danling Cao \\
  The Hong Kong Polytechnic University\\
     \And
  Hujun Yin \\
  The University of Manchester \\
}

\begin{document}
\maketitle

\begin{abstract}
Recovering 3D face models from 2D in-the-wild images has gained considerable attention in the computer vision community due to its wide range of potential applications. However, the lack of ground-truth labeled datasets and the complexity of real-world environments remain significant challenges. In this chapter, we propose a convolutional neural network-based approach, the Hierarchical Multi-Level Attention Network (MLANet), for reconstructing 3D face models from single in-the-wild images. Our model predicts detailed facial geometry, texture, pose, and illumination parameters from a single image.  Specifically, we employ a pre-trained hierarchical backbone network and introduce multi-level attention mechanisms at different stages of 2D face image feature extraction. A semi-supervised training strategy is employed, incorporating 3D Morphable Model (3DMM) parameters from publicly available datasets along with a differentiable renderer, enabling an end-to-end training process. Extensive experiments, including both comparative and ablation studies, were conducted on two benchmark datasets, AFLW2000-3D and MICC Florence, focusing on 3D face reconstruction and 3D face alignment tasks. The effectiveness of the proposed method was evaluated both quantitatively and qualitatively.
\end{abstract}


\section{Introduction}
Reconstructing 3D facial shape and texture from a single image remains a challenging task in computer vision, with applications in areas such as facial recognition, animation, and 3D avatar creation. While specialized hardware can achieve high-quality reconstructions, it often requires complex equipment and manual setup. Consequently, achieving robust, high-fidelity 3D face reconstruction from a single or sparse set of images is still a difficult problem.

A widely adopted approach to this task is the model-based method, which frames the reconstruction problem as a nonlinear optimisation or direct regression using CNNs. Many methods build on the 3D Morphable Model (3DMM) \cite{facemodel1999}, which models shape, expression, pose, and reflectance to manage data constraints and complex factors. However, 3DMM-based models relying on linear bases and limited datasets often lack the fidelity needed for capturing intricate facial details, particularly in challenging settings with varied lighting, large poses, or occlusions.

CNN-based architectures have been extensively used to predict 3DMM and related coefficients for 3D face reconstruction \cite{MoFA_Tewari_2017_ICCV,3DDFAv2, deep3d_deng, MGCNet, feng2018prn, depth2021robust}, but they face challenges in preserving both fine-grained details and broader facial structure. CNNs tend to blend depth and spatial features, resulting in low-level detail loss and occlusion artifacts. These models also lack adaptability to complex lighting and depend heavily on a single-scale representation, limiting their ability to capture spatial coherence and detail consistency.

To overcome these limitations, we introduce MLANet (Hierarchical Multi-Level Attention Network), a CNN framework for robust 3D face reconstruction from single in-the-wild images. MLANet incorporates a multi-stage attention framework that captures fine-grained local details while preserving global structural coherence. Specifically, it employs channel and spatial attention within backbone bottleneck layers to enhance critical facial regions and uses a context-aware fusion strategy to balance local and global information at fusion stages. This combined hierarchical and multi-stage attention approach enables MLANet to deliver high-quality reconstructions even under challenging conditions, such as occlusions, diverse lighting, and varying expressions.

In summary, the main contributions of this paper are:

\begin{enumerate}
    \item Hierarchical Multi-Level Attention Framework. We propose MLANet, a novel CNN-based framework that integrates hierarchical multi-level attention mechanisms at different stages of feature extraction, effectively capturing both fine-grained local details and maintaining global facial coherence. This approach enhances the accuracy and fidelity of 3D face reconstruction from single in-the-wild images.
    
    \item Context-Aware Feature Fusion with Attention Mechanisms. To address limitations in high-fidelity texture capture and spatial integrity, we introduce a context-aware fusion strategy. By incorporating attention mechanisms at the fusion levels, MLANet achieves a balanced integration of local and global information, significantly improving robustness across varied facial expressions, pose, and lighting conditions.
    
    \item We conduct comprehensive experiments, including both comparative and ablation studies, on the AFLW2000-3D and MICC Florence datasets. Our method demonstrates high performance in 3D face reconstruction and 3D face alignment tasks, highlighting both quantitative and qualitative robustness in real-world scenarios.
\end{enumerate}

\begin{figure}[t]
  \centering
  \includegraphics[width=1.0\textwidth]{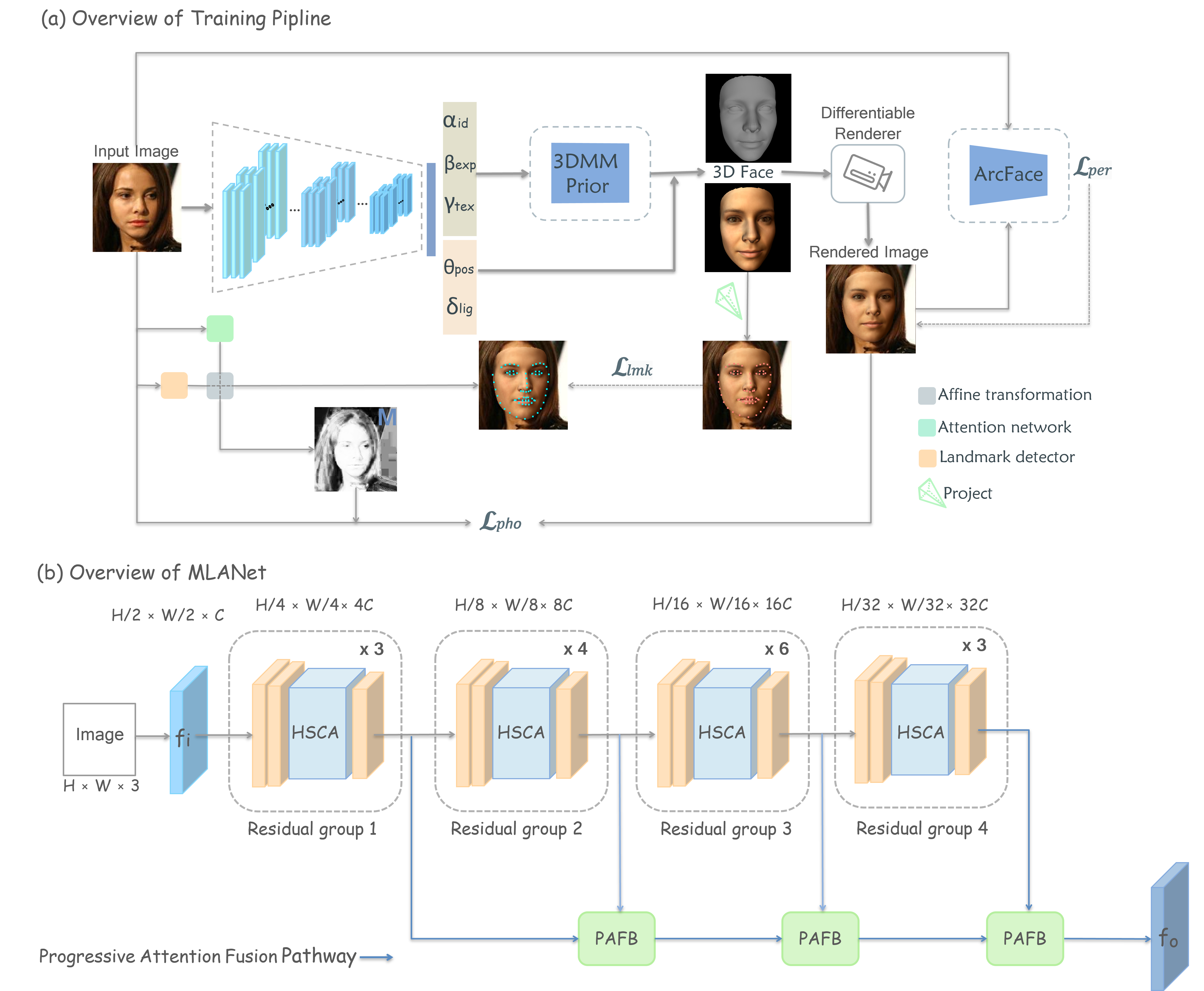}
  \caption[Overview of Proposed Hierarchical MLANet-based Framework]{Overview of the Proposed Hierarchical MLANet-based Framework: (a) \textit{Training Pipeline:} The input image is first subjected to affine transformation and then processed through an attention network to extract hierarchical features. A differentiable renderer is used to generate a 3D face, which is refined iteratively. (b) \textit{Overview of MLANet:} The input image is passed through four residual groups, each containing Hybrid Spatial-Channel Attention(HSCA) modules that extract multi-scale features. Progressive Attention Fusion Blocks (PAFB) integrate features across different scales to enhance feature representation.}
  \label{fig:MLAoverview}
\end{figure}

\section{Related Work}
\label{sec:headings}

\subsection{Statistical Morphable Model}
Introduced by \cite{facemodel1999}, the 3D Morphable Model (3DMM) \cite{survey} has long been a cornerstone in 3D face reconstruction, providing a statistical framework that encodes both the shape and texture of faces. The widely used Basel Face Model (BFM) \cite{bfm2009} facilitates the 3D face reconstruction process by parameterizing facial features through a linear combination of bases derived via Principal Component Analysis (PCA). This approach simplifies the complex 2D-to-3D reconstruction problem into a more manageable regression task, where the objective is to estimate the parameters of the 3DMM. Over the years, significant advancements have been made to enhance the original 3DMM framework, including models like FaceWarehouse \cite{facewarehouse2013}, Adaptive PCA-Model \cite{ietmaghari2014adaptive}, FLAME \cite{flame2017}, Surrey Face Model\cite{surrey2016}, and the LSFM \cite{lsfm2018},  which have aimed to enhance the versatility of the 3D Morphable Model.

\subsection{Optimization-based approaches}
Conventional methods for 3D face reconstruction using 3DMM relied on optimization techniques, which iteratively fit model parameters to a single input image by leveraging information such as image intensity, edges, and sparse facial landmarks \cite{Roth_2016_CVPR, yamaguchi2018high, Gecer_2019_CVPR, Yang_2020_CVPR, 9178990}. However, these optimization-based approaches can be computationally expensive and are often prone to local minima, particularly when handling pose variations, expressions, occlusions, or complex lighting conditions.

\subsection{Differentiable Renderers}
The advent of differentiable renderers, such as \cite{tfmeshrender_2018_CVPR, softrasterizer2019, KaolinLibrary2022,torch3d2020, nvdiffrast2020},  has significantly advanced 3D face reconstruction by enabling the transformation of a 3D face mesh into a 2D image. This process considers the face shape, texture, lighting, and other related factors, allowing the rendered image to be compared with the target image to compute image-level loss. Differentiable renderers have enriched supervised methods for 3D face reconstruction \cite{MoFA_Tewari_2017_ICCV, 3DDFA-v3} by making the rasterization process differentiable. This enables precise gradient computation for every pixel in the rendered outputs, thus enhancing the model's learning efficiency. The inclusion of a differentiable rendering layer ensures that reconstructed faces achieve both geometric accuracy and high visual fidelity, even in challenging scenarios with complex lighting, occlusions, and diverse facial expressions. Within our Hierarchical MLANet framework, we leverage the capabilities of Nvdiffrast \cite{nvdiffrast2020}. 

\subsection{Deep learning-based approaches}
Deep learning-based methods have revolutionized 3D face reconstruction, particularly in the direct regression of 3D Morphable Model (3DMM) coefficients from 2D images. This approach offers considerable improvements in speed and adaptability across various imaging conditions. Frameworks such as \cite{deep3d_deng, withoutsupervision2019, EMOCA2022, Li_2023_CVPR, 3DDFA-v3} using CNNs and \cite{Lin_2020_CVPR, Gao_2020_CVPR_Workshops, Lee_2020_CVPR, Qiu_2021_CVPR} utilizing GCNs effectively capture the underlying distribution of facial features, primarily focusing on the refinement of 3D facial shapes. These methods embed statistical parameters within a differentiable rendering framework, enabling robust end-to-end training and continuously refining the geometric accuracy of the reconstructed faces.

Many learning-based methods focus on refining the facial shape, a core aspect of 3D face reconstruction, by leveraging 3DMM. For instance, approaches like \cite{Tran_2017_CVPR, withoutsupervision2019, Yi_2019_CVPR, MICA2022} prioritize shape refinement to produce highly accurate facial geometry. In contrast, some methods aim to address both geometry and texture for a more comprehensive facial representation. While existing methods have explored various enhancements, these approaches often introduce additional complexities that may not directly focus on the specific goal of refining shape geometry.

  \subsection{Hierarchical Networks with Attention Mechanism}
Hierarchical network architectures have been pivotal in advancing computer vision, enabling the progressive extraction of features across multiple scales. Notable examples include ResNet \cite{resnet2016}, FPN \cite{fpn2017}, and variants of the Transformer \cite{ViTsurvey_hankai}, all of which excel in deep feature extraction and multi-scale or multi-stage processing. These architectures have been widely adopted in various computer vision tasks, including image super-resolution \cite{Imagesr2023}, object detection \cite{fpn2017}, and other visual tasks.The integration of attention mechanisms has significantly enhanced the effectiveness of hierarchical networks by enabling models to focus on the most relevant parts of the inputs. Attention mechanisms have been widely recognized for their impact across a range of tasks, such as image classification \cite{imgrecog2020, attimgcls2023}, image segmentation \cite{segattention2022, FsaNet2023}, and others.

Squeeze-and-excitation \cite{senet2018} introduced channel-wise attention to capture interdependencies between feature maps, while convolutional block attention module \cite{cbam2018} expanded this concept by adding spatial attention, thus enhancing context-awareness in feature extraction. Moreover, subsequent developments such as GENet \cite{hu2018genet} and TA \cite{misra2021TA} have introduced novel spatial attention mechanisms that further optimize feature representation more effectively.

\section{Methodology}\label{sec3}

In this section, we introduce the proposed framework for robust 3D face reconstruction from single in-the-wild images. We begin by providing a concise overview of the fundamental components and models that underpin our approach. This is followed by a detailed exposition of the multi-level attention mechanisms, which are applied at two critical stages: coarse feature extraction and contextual feature fusion. Finally, we present a comprehensive explanation of the training process, highlighting key procedures and design choices.

\subsection{Preliminaries}
This section outlines the fundamental components and models utilized in our work.

\subsubsection{Facial Geometry Prior.}
In a 3D morphable model for face modeling,  the shape and texture of a 3D face can be represented by the following formula:
\begin{equation}
    \begin{aligned}&\mathbf{S}(\boldsymbol{\alpha}_{id},\boldsymbol{\beta}_{exp})=\overline{\mathbf{S}}+\mathbf{A}_{id}\boldsymbol{\alpha}_{id}+\mathbf{A}_{exp}\boldsymbol{\beta}_{exp}\\&\mathbf{T}(\boldsymbol{\gamma}_{tex})=\overline{\mathbf{T}}+\mathbf{A}_{tex}\boldsymbol{\gamma}_{tex}\end{aligned}
\end{equation}

where $\overline{\mathbf{S}}$ and $\overline{\mathbf{T}}$ are the mean shape and texture, $\mathbf{A}_{id}$, $\mathbf{A}_{exp}$, and $\mathbf{A}_{tex}$ are the shape bases, expression bases, and texture bases, respectively. $\boldsymbol{\alpha}_{id}\in\mathbb{R}^{80}$, $\boldsymbol{\beta}_{exp}\in\mathbb{R}^{64}$, and $\boldsymbol{\gamma}_{tex}\in\mathbb{R}^{80}$ are the coefficients of the shape, expression, and texture, respectively. We employ the widely-used 2009 Basel Face Model(BFM) as the underlying 3DMM for $\overline{\mathbf{S}}$, $\overline{\mathbf{T}}$, $\mathbf{A}_{id}$, and $\mathbf{A}_{tex}$, excluding the ear and neck regions. For the expression model $\mathbf{A}_{exp}$, we adopt the bases built from the FaceWarehouse dataset, utilizing delta blendshapes to represent facial expressions.

\subsubsection{Pose \& Camera Model.}
In our framework, we adopt a weak perspective projection to model the transformation from 3D face geometry to 2D image space. The 3D face shape, denoted as $\mathbf{S}$, is projected onto the 2D plane, yielding the projected geometry $\mathbf{V_{2d}}$:
\begin{equation}
    \mathbf{V_{2d}} = \mathbf{P_r} * (\mathbf{R*S+t})
\end{equation}
where $\mathbf{P_r}$ is the projection matrix, and $\mathbf{S}$ is the pose-independent face shape. $\mathbf{R}$ and $\mathbf{t}$ are the 3D face rotation matrix and translation matrix, respectively.We implement a perspective camera model with an empirically determined focal length for the 3D-to-2D projection geometry. Therefore, the 3D face pose $\boldsymbol{\theta}_{pos}$ is parameterized by an Euler rotation $\mathbf{R} \in \mathbf{SO(3)}$ and translation $t \in \mathbf{R}^3$.

\subsubsection{Illumination Model.}
To achieve realistic face rendering, we approximate scene illumination using Spherical Harmonics(SH) following \cite{SH2001}. Similar to \cite{deep3d_deng}, we choose the first three bands of SH basis functions, and using the face texture and surface normal as input, then the shaded texture $\mathbf{T_{sh}}$ is computed as follows:
\begin{equation}
    \mathbf{T_{sh}} = \mathbf{T}(\boldsymbol{\gamma}_{tex}) \odot \sum_{k=1}^9 \boldsymbol{\delta}_{lig} \Psi_k(\mathbf{n})
\end{equation}

where \( \odot \) denotes the Hadamard product, $\mathbf{n}$ represents the surface normals of the 3D face model, \( \Psi_k : \mathbb{R}^3 \to \mathbb{R} \) is the SH basis function, and $\boldsymbol{\delta}_{lig} \in \mathbf{R}^9$ is the corresponding SH parameters.

\subsubsection{Differentiable Rendering Layer.}
In summary, the vector $\mathbf{x}=[\boldsymbol{\alpha}_{id}, \boldsymbol{\beta}_{exp}, \boldsymbol{\gamma}_{tex}, \boldsymbol{\theta}_{pos}, \boldsymbol{\delta}_{lig}]$ represents the undetermined coefficients. These coefficients are used to reconstruct the 3D face and render it to the 2D image plane. We define the differentiable rendering layer using Nvdiffrast \cite{nvdiffrast2020} to compute the rendered face. 

\subsection{Overview of Hierarchical MLANet}
In this section, we present the proposed framework, emphasizing the multi-level attention mechanisms integrated at different granularity levels. An overview of our multi-level attention network structure can be found in Figure~\ref{fig:MLAoverview}.

\subsubsection{Hierarchical Attention-Based Feature Extraction}
The MLANet architecture is designed to extract hierarchical features from 2D facial images efficiently, leveraging both the 3D Morphable Model (3DMM) and rendering information. The backbone of our network is based on ResNet50 \cite{resnet2016}. As depicted in Figure \ref{fig:MLAoverview}, the input image is first processed through initial convolutional layers, producing feature maps at four different scales \(\left\{\frac{1}{4}, \frac{1}{8}, \frac{1}{16}, \frac{1}{32}\right\}\) across four stages. Each stage consists of residual blocks with integrated Hybrid Spatial-Channel Attention(HSCA) modules, which effectively encode both spatial and channel information. These HSCA modules enhance the network's ability to capture essential features for accurate 3D face reconstruction.

Following the extraction of features at each stage, the network further refines these features through additional convolutional layers and residual blocks. At the end of each stage, Progressive Attention Fusion Blocks(PAFB) are introduced to combine global and local features. These blocks enhance the overall feature set by integrating global attention to capture overall facial structure and local attention to focus on detailed facial parts.

\subsubsection{Hybrid Spatial-Channel Attention Module (HSCA)}
The HSCA module in the Hierarchical MLANet framework employs both spatial and channel attention mechanisms to enhance the representation of facial features from the extracted feature maps. Integrated immediately following each \( 3 \times 3 \) convolution layer in the ResNet bottleneck, the HSCA module processes feature tensors \( X \in \mathbb{R}^{H \times W \times C} \), where \( H \), \( W \), and \( C \) denote the spatial height, width, and channel dimensions, respectively. This module first applies spatial attention to focus on spatially informative regions within the feature map, followed by channel attention to prioritize essential channels based on their relevance to facial structures.  As illustrated in Figure \ref{fig:MLAHSCA},  the HSCA module is selectively integrated within specific bottleneck layers of the network to optimize feature extraction.

    \begin{figure}
      \centering
      \includegraphics[width=1.0\columnwidth]{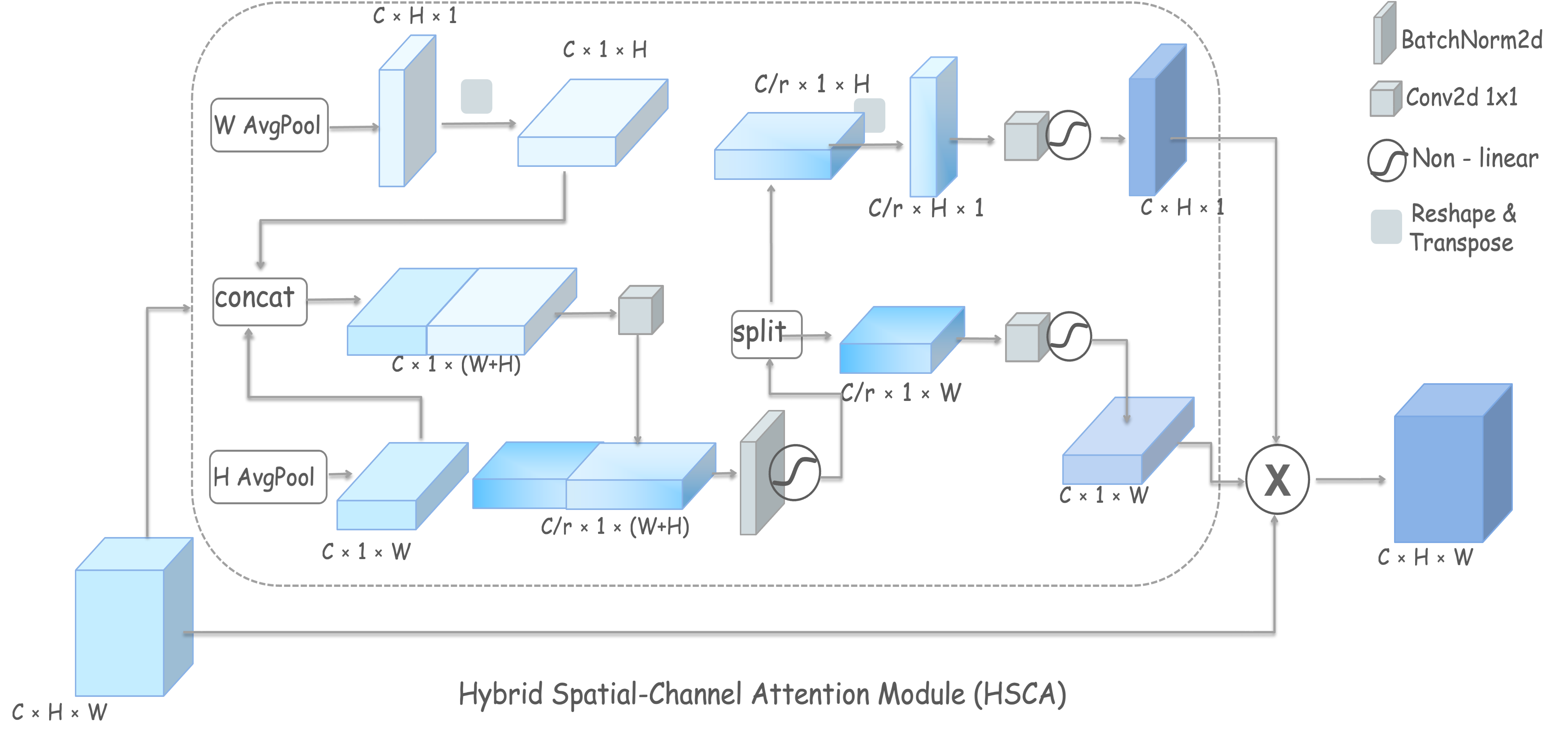}
      \caption[Overview of Hybrid Spatial-Channel Attention Module (HSCA)]{Hybrid Spatial-Channel Attention Module (HSCA). The HSCA module integrates spatial and channel attention mechanisms. The input feature map undergoes average pooling along width and height to produce spatial attention maps. These are concatenated, reshaped, and processed to form channel attention maps, which are split and applied to the feature map. The final output is a refined feature map that combines both spatial and channel attention.}
      \label{MLAHSCA}
    \end{figure} 
In the spatial attention branch, the input feature map $\mathbf{F}_{i} \in \mathbb{R}^{C \times H \times W}$ undergoes horizontal and vertical attentions. Horizontal attention is achieved by average pooling along the width axis, producing a feature representation of dimension \( \mathbb{R}^{C \times H \times 1} \). Similarly, vertical attention involves average pooling along the height axis, yielding a feature representation of dimension \( \mathbb{R}^{C \times 1 \times W} \). These representations are then concatenated and passed through a convolutional layer, generating spatial attention maps that emphasize significant spatial locations through weighting adhustments:
\begin{equation}
\small
    \mathbf{F_{spatial}} = \boldsymbol{\sigma} \left( \text{Conv} \left( [\text{AvgPool}_H(\mathbf{F}); \text{AvgPool}_W(\mathbf{F})] \right) \right) \cdot \mathbf{F} 
\end{equation}
where $\boldsymbol{\sigma}$ denotes the Non-linear function, and $[\boldsymbol{\cdot} \,;\, \boldsymbol{\cdot}]$ represents concatenation.

In the channel attention branch, the reweighted feature map $\mathbf{F_{spatial}}$ undergoes channel attention. The feature map is average-pooled along the spatial dimensions to generate a channel descriptor of size $\mathbb{R}^{C \times 1 \times 1}$. This descriptor is passed through a convolutional layer followed by sigmoid activation to produce channel attention maps. The channel attention maps adjust the importance of each channel, refining the feature representation by focusing on significant channels.
\begin{equation}
\small
    \mathbf{F_{channel}} = \boldsymbol{\sigma} \left( \text{Conv} \left( \text{AvgPool}(\mathbf{F_{spatial}}) \right) \right) \cdot \mathbf{F_{spatial}}
\end{equation}

The final output of the HSCA module combines both spatial and channel attention mechanisms, forming a comprehensive hybrid attention map that enhances the network's ability to focus on long-range dependencies in one spatial direction, as well as significant channels and relevant spatial locations. This dual focus sharpens the accuracy in capturing essential facial details and suppresses background noise, providing more precise and useful facial shape features for the resulting 3D face model.

\begin{figure}[t]
  \centering
    \includegraphics[width=1.0\columnwidth]{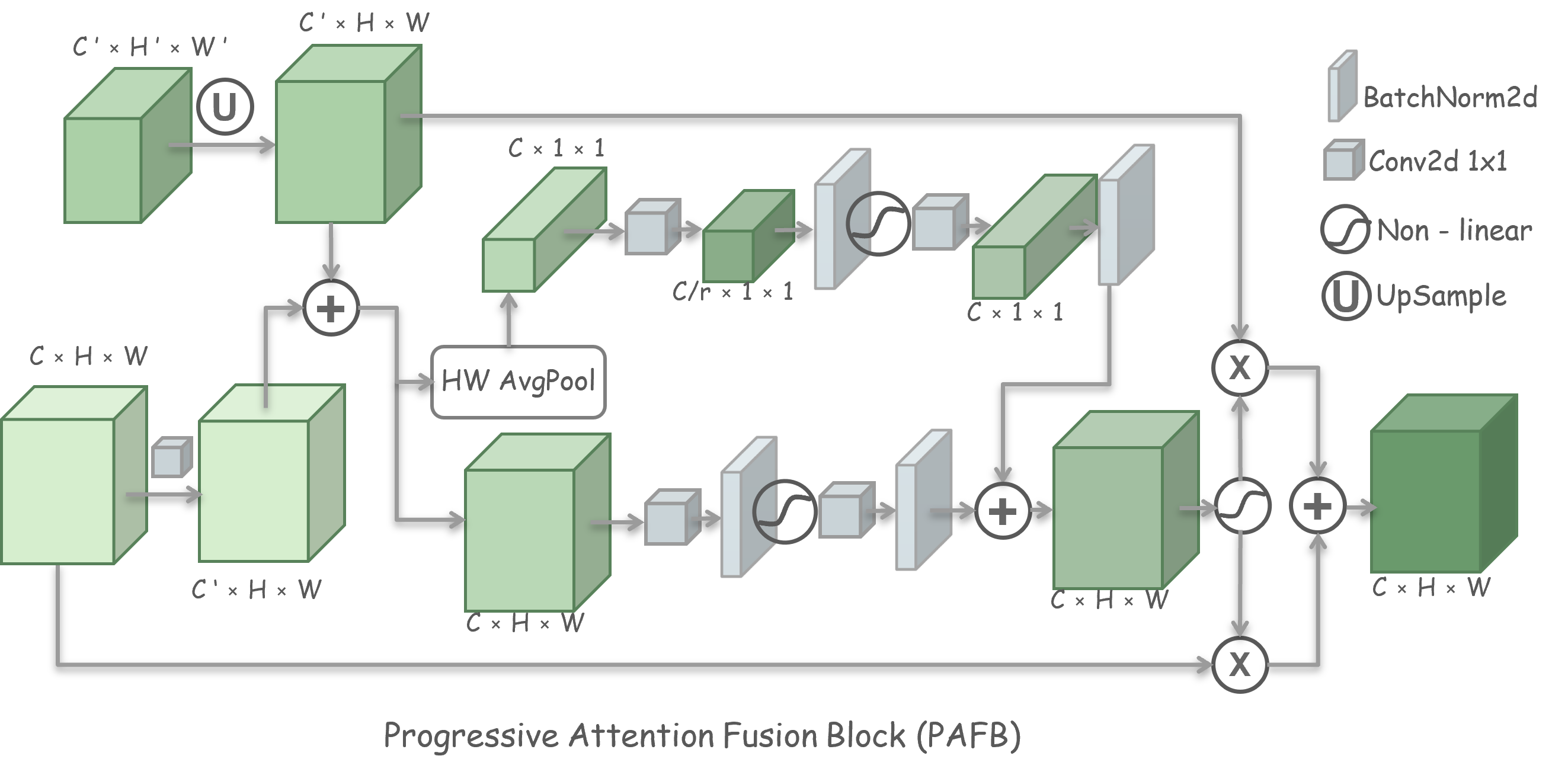}
  \caption[Overview of Progressive Attention Fusion Block (PAFB)]{Progressive Attention Fusion Block (PAFB). The PAFB integrates multi-scale features by combining high-resolution feature maps with upsampled low-resolution maps. This process involves spatial and channel attention mechanisms, enhancing the feature representation by emphasizing important spatial locations and channels. The refined feature maps are then fused through element-wise operations for further processing.}
  \label{MLAPAF}
\end{figure}
\subsubsection{Progressive Attention Fusion Module (PAF)}
In the Hierarchical MLANet framework, Progressive Attention Fusion (PAF) module, based on the AFF mechanism \cite{AFF2021}, are strategically integrated at the end of each stage to effectively combine hierarchical features extracted by the regressor at four different spatial resolutions \(\left\{\frac{1}{4}, \frac{1}{8}, \frac{1}{16}, \frac{1}{32}\right\}\). These PAF modules bridge the semantic gap between high-resolution, low-level detailed features and low-resolution, high-level abstract features, progressively enhancing feature fusion.

As illustrated in Figure~\ref{MLAPAF}, the Progressive Attention Fusion Block (PAFB) integrates multi-scale feature maps by progressively downsampling and aligning high-resolution feature maps to match the resolution and channel dimensions of the next lower-resolution feature map. Specifically, the high-resolution feature map \(\mathbf{F}_{\text{high}} \in \mathbb{R}^{C \times H \times W}\) is downsampled via a $1 \times 1$ convolution with a stride of 2, reducing its spatial dimensions to \(\frac{H}{2} \times \frac{W}{2}\) while doubling the number of channels to \(2C\). This ensures that the downsampled high-resolution feature map aligns with the dimensions of the low-resolution feature map \(\mathbf{F}_{\text{low}} \in \mathbb{R}^{2C \times \frac{H}{2} \times \frac{W}{2}}\) from the subsequent stage. The downsampled feature map \(\text{D}(\mathbf{F}_{\text{high}})\) is then fused with \(\mathbf{F}_{\text{low}}\) using an attention fusion mechanism, which adjusts both spatial and channel information. The fusion is performed element-wise, allowing for the preservation of both global and local context across scales. This process is repeated at multiple levels of the network, progressively combining features from various stages to produce a more robust feature representation.
The fusion process is described by the following equation:
\begin{equation}
    \mathbf{F}_{\text{combined}} = \text{Concat}(\mathbf{F}_{\text{low}}, \mathbf{D}(\mathbf{F}_{\text{high}}))
\end{equation}
The combined feature map \(\mathbf{F}_{\text{combined}}\) is subsequently processed through a dual-attention mechanism, which integrates both global and local attention branches. This approach is designed to enhance the model’s ability to capture context at multiple levels, thereby improving feature representation across spatial and channel dimensions.

In the global attention branch, global contextual information is captured by applying global average pooling (GAP) to the combined feature map, reducing the spatial dimensions to $1 \times 1$ for each channel. This pooling operation summarizes the global context of the entire feature map, creating a compact global representation. The resulting global context vector is then passed through a series of $1 \times 1$ convolutional layers followed by batch normalization and ReLU activation. These layers act as fully connected layers that model dependencies across channels based on the global context.

Simultaneously, in the local attention branch, a series of $1 \times 1$ convolutions are applied to the combined feature map without altering its spatial resolution, thereby preserving fine-grained local dependencies. These two attention outputs are combined to create a refined feature representation \(\mathbf{F}_{\text{refined}}\):
\begin{equation}
     \begin{aligned}&\mathbf{F}_{\text{GlobalAtt}} = \text{FC}(\text{GAP}(\mathbf{F}_{\text{combined}}))\\&\mathbf{F}_{\text{LocalAtt}} = \text{FC}(\mathbf{F}_{\text{combined}})
     \\&\mathbf{F}_{\text{refined}} = \text{LocalAtt}(\mathbf{F}_{\text{combined}}) + \text{GlobalAtt}(\mathbf{F}_{\text{combined}})\end{aligned}
\end{equation}
Next, the refined feature map \(\mathbf{F}_{\text{refined}}\) is used to compute the attention weights \(\omega_{1}\) and \(\omega_{2}\) via a sigmoid activation function \(\sigma\), which are applied to the original high-resolution and low-resolution feature maps, respectively, and these attention weights are then used to modulate the contribution of the high-resolution and low-resolution feature maps in the final fused representation:
\begin{equation}
    \begin{aligned}&\omega_{1} = \sigma\mathbf{F}_{\text{refined}}
    \\& \omega_{1} + \omega_{2} = 1
    \\&\mathbf{F}_{\text{final}} = \omega_{1} \odot \mathbf{F}_{\text{high}}  + \omega_{2} \odot \mathbf{F}_{\text{low}}\end{aligned}
\end{equation}
where \(\odot\) denotes element-wise multiplication. This weighted fusion mechanism allows the model to dynamically prioritize important features across scales, ensuring that the representation captures both global and local context effectively. The final output \(\mathbf{F}_{\text{final}}\) is then passed through the subsequent stages of the framework, facilitating improved reconstruction of detailed and abstract shape information. 

\subsection{Training Process}
To reconstruct a 3D face from image $\mathbf{I}$, the overall training objective is formulated by combining multiple loss terms, each focusing on different aspects of the reconstruction task. The total loss \( \mathcal{L} \) is defined as follows:

\begin{equation}
\begin{aligned}
    \mathcal{L} = \lambda_{\text{pho}} \mathcal{L}_{\text{pho}} + \lambda_{\text{per}} \mathcal{L}_{\text{per}} + \lambda_{\text{lmk}} \mathcal{L}_{\text{lmk}} \\
    + \lambda_{\text{3dmm}} \mathcal{L}_{\text{3dmm}} + \lambda_{\text{refl}} \mathcal{L}_{\text{refl}}
\end{aligned}
\end{equation}

where the $\mathcal{L}_{\text{pho}}$ is the photometric loss, $\mathcal{L}_{\text{per}}$ is the perceptual loss, $\mathcal{L}_{\text{lmk}}$ is the landmark reprojection loss, $\mathcal{L}_{\text{3dmm}}$ and $\mathcal{L}_{\text{refl}}$ is the 3dmm coefficients regularization loss, and $\mathcal{L}_{\text{refl}}$ is the reflectance loss. The balance weights $\lambda_{\text{pho}}$, $\lambda_{\text{per}}$, $\lambda_{\text{lmk}}$, $\lambda_{\text{3dmm}}$, and $\lambda_{\text{refl}}$ are set to 1.9, 0.2, 1.6e-3, 3e-4, and 4.5, respectively.

\subsubsection{Photometric Loss.}
Inspired by \cite{deep3d_deng}, we utilize the differentiable renderer \cite{nvdiffrast2020} to obtain the rendered image $mathbf{I}^{r}$. We use the $l_{2}$ loss to compute the photometric discrepancy between the input image $\mathbf{I}$ and the rendered image $\mathbf{I^{r}}$. The photometric loss is defined as
\begin{equation}
    \mathcal{L}_{\text{pho}} = \frac{\sum_{i \in \mathcal{M}} A_i \cdot \left\| \mathbf{I} - \mathbf{I}^r \right\|_2}{\sum_{i \in \mathcal{M}} A_i},
\end{equation}
where i represents the pixel index, and M denotes the reprojected face region generated by the differentiable renderer. A is a face mask with a value of 1 in the face skin region and a value of 0 elsewhere obtained by an existing face segmentation method \cite{saito2016real}, which is capable of reducing errors caused by occlusion, such as eyeglasses.

\subsubsection{Perceptual Loss.}
We employ the ArcFace model \cite{arcface} for feature embedding to compute the perceptual loss. The perceptual loss is defined as:

\begin{equation}
    \mathcal{L}_{\text{per}} = 1 - \frac{\Gamma(\mathbf{I}) \cdot \Gamma(\mathbf{I}^r)}{\|\Gamma(\mathbf{I})\|_2 \cdot \|\Gamma(\mathbf{I}^r)\|_2},
\end{equation}

where $\Gamma(\cdot)$ denotes the feature embedding obtained from the ArcFace model, $\mathbf{I}$ is the input image, and $\mathbf{I}^r$ is the rendered image. The loss measures the cosine similarity between the feature embeddings of the input and rendered images.

\subsubsection{Landmark Rejection Loss.}
We utilize 68 facial landmarks \cite{ietland2019survey}, automatically detected from the input image, as weak supervision. For each training image, the face alignment network \cite{bulat2017far} detects the 2D positions of these landmarks, denoted as $\{\mathbf{p}_n\}$. To calculate the reprojection loss during training, we project the 3D landmarks of the reconstructed face shapes onto the 2D image plane to obtain the corresponding 2D landmarks, denoted as $\{\mathbf{p}'_n\}$. The reprojection loss is calculated as follows:

\begin{equation}
    \mathcal{L}_{\text{lmk}} = \frac{1}{N} \sum_{n=1}^{N} \omega_n \left\| \mathbf{p}_n - \mathbf{p}'_n \right\|_2^2,
\end{equation}

where \( N \) is the number of facial landmarks (68), and \( \omega_n \) is the weight assigned to the \( n \)-th landmark, which we set to 20 for the inner mouth and 1 for the others \cite{deep3d_deng}. 

\subsubsection{Regularization Loss.}
In order to preserve the integrity of facial shape and texture, we employ a regularization loss on the estimated 3DMM coefficients to enforce a prior distribution towards the mean face. The coefficient loss is defined as:

\begin{equation}
    \mathcal{L}_{\text{3DMM}}(\mathbf{x}) = \lambda_{\alpha} \left\| \alpha_{id} \right\|_2^2 + \lambda_{\beta} \left\| \beta_{exp} \right\|_2^2 + \lambda_{\gamma} \left\| \gamma_{tex} \right\|_2^2
\end{equation}

Following \cite{deep3d_deng}, we set the weights \( \lambda_{\alpha} = 1.0 \), \( \lambda_{\beta} = 0.8 \), and \( \lambda_{\gamma} = 1.7e-2 \), and through the reflectance loss to favor a constant skin albedo similar to \cite{tewari2018self} . Which is formulated as:

\begin{equation}
    \mathcal{L}_{\text{refl}} = \frac{\sum (\mathbf{M} \odot (\mathbf{T} - \overline{\mathbf{T}}))^2}{\sum \mathbf{M}}
\end{equation}

where \(\overline{\mathbf{T}} = \frac{\sum (\mathbf{M} \odot \mathbf{T})}{\sum \mathbf{M}}\) is the mean texture within the masked region, \(\mathbf{M}\) is a binary mask indicating the face region, and \(\odot\) denotes element-wise multiplication. 

The overall regularization loss combines these two components:

\begin{equation}
    \mathcal{L}_{\text{reg}} = \lambda_{\text{3dmm}} \mathcal{L}_{\text{3dmm}} + \lambda_{\text{refl}} \mathcal{L}_{\text{refl}}
\end{equation}

where $\lambda_{\text{3dmm}}=3e-4$ and $\lambda_{\text{refl}} = 4.5$ are the balance weights between the 3DMM coefficients loss and the reflectance loss.

\section{Experiments and Results}
    \subsection{Reconstruction Frameworks}
    We implement Hierarchical MLANet based on PyTorch and utilize the differentiable renderer from Nvdiffrast \cite{nvdiffrast2020}. We employ ResNet-50 \cite{resnet2016} as the initial backbone. The output dimension of the last fully connected layer is modified to 257 neurons to output the required coefficients. Input images are detected, cropped, and aligned by \cite{mtcnn}, resized to 224 × 224.
    \begin{table}[t]
      \centering
      \caption[Comparison of NME (\%) for Face Alignment with 68 Landmarks on AFLW2000-3D Dataset across Different Yaw Intervals]{Comparison of NME (\%) for Face Alignment Results on the AFLW2000-3D Dataset across Different Yaw Intervals}
      \label{tab:NME_AFLW}
      \begin{tabular}{lcccc}
        \toprule
        \multirow{2}{*}{\textbf{Method}} & \multicolumn{4}{c}{\textbf{AFLW2000-3D (68 pts)}}\\
        \cline{2-5}
        & [0°, 30°] & [30°, 60°] & [60°, 90°] & \textbf{Mean}\\
	       \toprule
        SDM 
        & 3.67 & 4.94 & 9.67 & 6.12\\
        3DDFA 
        & 3.78 & 4.54 & 7.93 & 5.42 \\
        3DDFA+SDM 
        & 3.43 & 4.24 & 7.17 & 4.94 \\
        PRNet 
        & 2.75 & 3.51 & 4.61 & 3.62\\   
        3DDFAv2 
        & 2.63 & 3.42 & 4.48 & 3.51 \\
        Yu \textit{et al.} 
        & \textbf{2.56} & \textbf{3.11} & \underline{4.45} & \textbf{3.37} \\
        MLANet (ours)  & \underline{2.61} & \underline{3.38} & \textbf{4.41} &  \underline{3.39}\\
        \bottomrule
      \end{tabular}
    \end{table} 
    \subsection{Datasets and Evaluation Metrics}
We employ publicly available datasets for training, including CelebA \cite{celebA} and 300W-LP \cite{3DDFA}, for self-supervised learning. Additionally, face pose augmentation is applied using the method from \cite{skindetection, deep3d_deng}. After merging these two in-the-wild datasets, we employ a face alignment network (FAN) \cite{bulat2017far} to extract 68 2D facial landmarks from each image, which serve as the ground truth. Images where landmarks were not successfully detected are discarded, resulting in a final dataset of approximately 45K images as our training set. Furthermore, during training, we augment the input images by applying random horizontal shifts with a 50\% probability, as well as random rotations.

To evaluate the performance of our model, we assess it on two primary tasks: 3D face reconstruction and dense face alignment. For face alignment, we evaluate accuracy across different yaw angles (\(\psi\)) using the AFLW2000-3D \cite{ALFW-3D} dataset. We compute the normalized mean error (NME) for three yaw angle intervals: \( 0^\circ \leq \psi < 30^\circ \), \( 30^\circ \leq \psi < 60^\circ \), and \( 60^\circ \leq \psi \leq 90^\circ \). 

For 3D face reconstruction, we compute the geometric reconstruction error using the point-to-plane Root Mean Square Error (RMSE) on the MICC Florence \cite{Florence} dataset. This metric quantifies the geometric discrepancy between the predicted 3D face shapes and the ground truth by measuring the perpendicular distance from each predicted point to the corresponding surface in the ground truth. We provide both qualitative and quantitative comparisons with previous works to demonstrate the robustness and accuracy of our approach in reconstructing 3D facial geometry.

\begin{table}[h]
\centering
\setlength{\tabcolsep}{10pt} 
\renewcommand{\arraystretch}{1.2}
\caption[Comparison of Mean Root Mean Squared Error (RMSE) for Face Reconstruction on the MICC Florence Dataset]{Comparison of Mean Root Mean Squared Error (RMSE) for Face Reconstruction on the MICC Florence Dataset (in mm). Results are shown as Mean $\pm$ Standard Deviation.}
\label{tab:RMSE_MICC}
\begin{tabular}{lccc}
\toprule
\multirow{2}{*}{\textbf{Method}} & \multicolumn{3}{c}{\textbf{MICC Florence}}\\
\cline{2-4}
& Cooperative & Indoor & Outdoor\\
\toprule
RingNet 
& $2.09 \pm 0.48$ & $2.13 \pm 0.46$ & $2.10 \pm 0.47$ \\
CPEM 
& $2.08 \pm 0.59$ & 2.09 $\pm$ 0.54 & $2.02 \pm 0.54$ \\
Yu \textit{et al.} 
& $1.74 \pm 0.58$ & $1.75 \pm 0.49$ & $1.79 \pm 0.46$ \\
Yu \textit{et al.} (Pytorch) 
& $1.67 \pm 0.49$ & 1.68 $\pm$ 0.51 & $1.71 \pm 0.53$ \\
Tran \textit{et al.} 
& $1.97 \pm 0.49$ & $2.03 \pm 0.45$ & $1.93 \pm 0.49$ \\
MGCNet \textit{et al.} 
& $1.78 \pm 0.55$ & 1.78 $\pm$ 0.54 & $1.81 \pm 0.59$  \\
Booth \textit{et al.} 
& $1.82 \pm 0.29$ & 1.85 $\pm$ 0.22 & \textbf{1.63 $\pm$ 0.16}  \\
3DDFA-v2 
& \textbf{1.65 $\pm$ 0.56} & \underline{$1.66 \pm 0.50$} & 1.71 $\pm$ 0.66 \\
Ours (MLANet)  & \underline{$1.66 \pm 0.51$} & \textbf{1.66 $\pm$ 0.49} & \underline{1.66 $\pm$ 0.53}  \\
\bottomrule
\end{tabular}
\end{table}

    \subsection{Implementation Details}
    During the training process, we use the pre-trained ResNet-50 on ImageNet \cite{resnet2015imagenet} as initialization, we change the output dimension of the
last fully connected layer to output all coefficients and adopted Adam \cite{adam} as the optimizer with an initial learning rate $lr = 1e-4$ and then decayed by a factor of 0.1 for every 10 epochs. We train our network with a batch size of 32 on a single NVIDIA Tesla V100 GPU.

\begin{table}[t]
    \centering
    \caption[Ablation Study on AFLW2000-3D and MICC Florence Dataset]{Ablation Study on AFLW2000-3D Dataset. The “HSCA” denotes using the spatial-channel attention module or not, and the “PAF” denotes using the attention fusion module or not.}
    \label{tab:ablation}
    \begin{tabular}{lcc}
    \toprule
    \textbf{Model} & \textbf{NME on AFLW2000-3D} & \textbf{RMSE on MICC} \\
    \midrule
    our model w/o HSCA, PAF & 3.62 & $1.68 \pm 0.54$ \\
    our model w/o HSCA & 3.51 & 1.67 $\pm$ 0.45 \\
    our model w/o PAF & 3.56 & 1.67 $\pm$ 0.58 \\
    our full model & \textbf{3.41} & \textbf{1.66 $\pm$ 0.50} \\
    \bottomrule
    \end{tabular}
\end{table}

\subsection{Evaluation on Face Alignment}
To quantitatively evaluate the face alignment accuracy of our proposed MLANet, we utilize the Normalized Mean Error (NME), a widely adopted metric for assessing alignment precision. The NME is computed on the AFLW2000-3D benchmark and measures the normalized mean Euclidean distance between corresponding landmarks in the predicted result $p_{i}$ and the ground truth $p'_{i}$. The calculation is formalized as follows:
\begin{equation}
    \text{NME} = \frac{1}{N} \sum_{i=1}^{N} \frac{\lVert p_i - p'_i \rVert_2}{d}
\end{equation}
where $N$ denotes the number of landmarks, $\lVert \cdot \rVert_2$ is the Euclidean distance, and $d$ is a normalization factor defined as \( \sqrt{h \ast w} \) where $h$ and $w$ are the hight and width of the bounding box around the ground truth landmarks.

To provide a thorough evaluation, we follow the settings in previous studies \cite{feng2018prn, 3DDFAv2, MGCNet, SADRNET} and categorize the AFLW2000-3D dataset into yaw angle ranges of $[ 0^\circ,  30^\circ)$, $[ 30^\circ,  60^\circ)$, and $[ 60^\circ,  90^\circ]$. Within each range, balanced subsets are used to ensure fair comparisons across different head poses.

\begin{figure}[!h]
  \centering
  \includegraphics[width=0.9\textwidth]{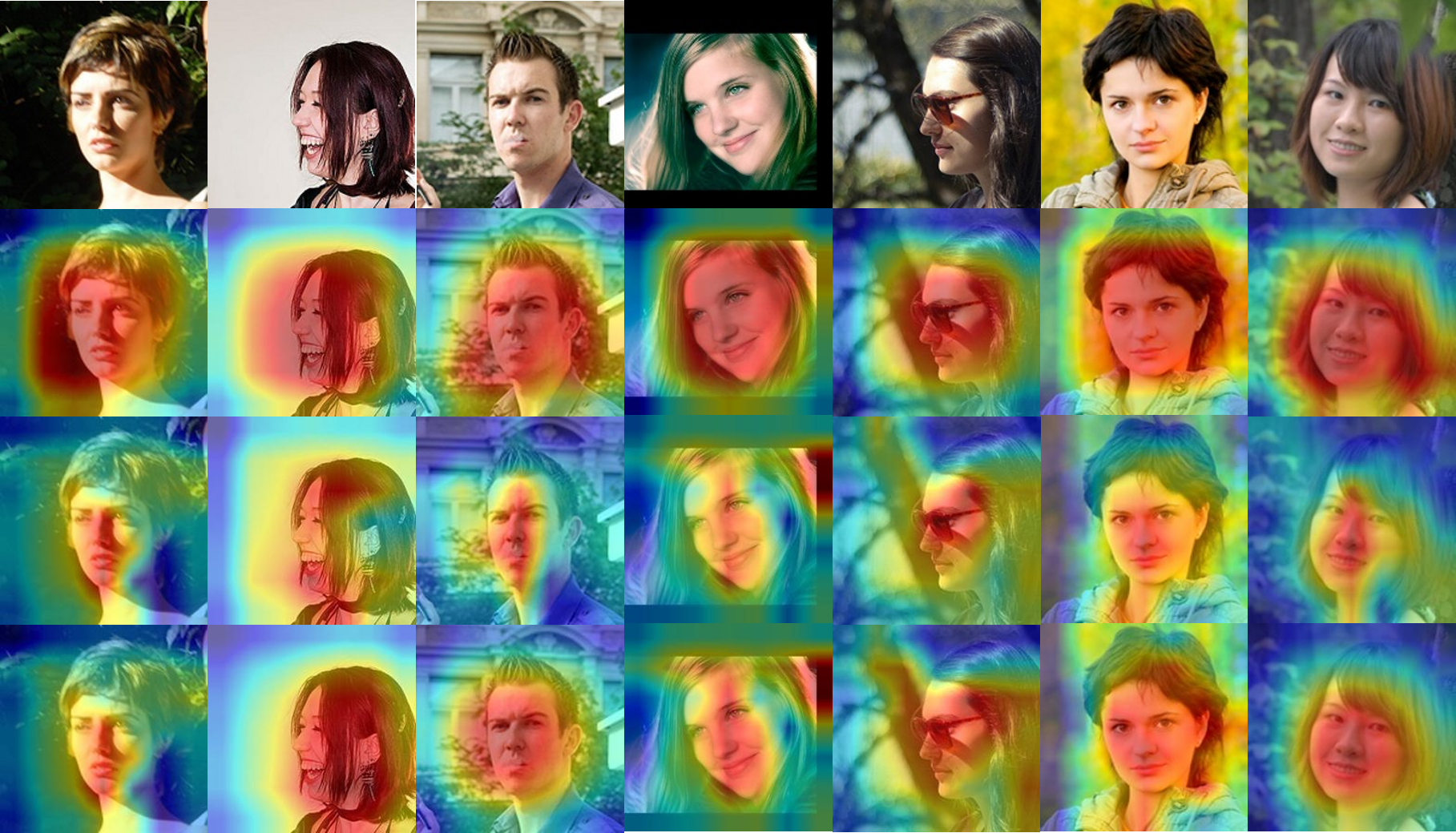}
  \caption[Visualization of Feature Maps with Multi-Level Attention Modules]{Visualization of the learned feature maps on images from the AFLW2000-3D dataset. The first row displays the original input images. The second row shows feature maps generated by ResNet50. The third row presents feature maps produced by the HSCA module, and the fourth row shows feature maps generated by the combined HSCA and PAF modules in our method.}
  \label{fig:grad_cam}
\end{figure}

The results are presented in Table~\ref{tab:NME_AFLW}, where our method is compared with existing approaches, including SDM \cite{SDM2015}, 3DDFA \cite{3DDFA}, PRNet \cite{feng2018prn}, 3DDFA-V2 \cite{3DDFAv2}, and Deep3D \cite{deep3d_deng}. The reported results for the comparison methods are sourced from their respective publications.

As shown in the Table~\ref{tab:NME_AFLW}., MLANet achieves competitive performance, particularly in the  $[ 30^\circ,  60^\circ)$, $[ 60^\circ,  90^\circ]$ yaw intervals, and in the overall mean NME. Our method achieves results on par with the methods SADRNet, 3DDFA-v2, and outperforms other methods in most categories. 

\subsection{Evaluation of Face Reconstruction}
We evaluate the 3D face reconstruction performance using the point-to-plane Root Mean Square Error (RMSE). For comparison, we include results from RingNet \cite{withoutsupervision2019}, CPEM \cite{CPEM}, Yu \textit{et al.} \cite{deep3d_deng}, Tran \textit{et al.} \cite{Tran_2017_CVPR}, 3DDFA-V2 \cite{3DDFAv2}, MGCNet \cite{MGCNet}, and Booth \textit{et al.} \cite{booth20173d} on the MICC Florence dataset. This dataset includes 53 subjects, each with a high-quality neutral face scan and three video clips captured under different environmental conditions (cooperative, indoor, and outdoor) of increasing difficulty.

Following the methodology described in \cite{tfmeshrender_2018_CVPR}, we compute the point-to-plane RMSE by averaging errors across frames within each scenario (cooperative, indoor, outdoor) and then averaging across all three scenarios. To ensure consistent error measurement, we align the reconstructed 3D faces to ground truth meshes cropped to a 95mm radius around the nose tip, as proposed in \cite{genova2018unsupervised, deep3d_deng, Gecer_2019_CVPR}. The alignment is performed using the Iterative Closest Point (ICP) algorithm \cite{ICP} with isotropic scaling to reduce discrepancies, after which we compute the point-to-plane distances between the aligned meshes.

The results are presented in Table~\ref{tab:RMSE_MICC}. The values for Booth \textit{et al.} are derived from the GANFIT study \cite{Gecer_2019_CVPR}, while the results for RingNet and CPEM are sourced from the CPEM paper \cite{CPEM}. The performance of MGCNet is obtained from Qixin \textit{et al.} \cite{MM}, and the results for Tran \textit{et al.} \cite{Tran_2017_CVPR} are taken from Yu \textit{et al.} \cite{deep3d_deng}. The performance of Yu \textit{et al.} \cite{deep3d_deng}, Yu \textit{et al.} (PyTorch) \cite{deep3d_deng}, and 3DDFA-v2 \cite{3DDFAv2} were computed by running their pre-trained models using the methodology described above. From the results, it can be seen that our method achieves competitive reconstruction accuracy, with smaller RMSE values compared to previous works in most conditions. Our method outperforms other methods except for 3DDFA-v2 in the cooperative setting and Booth \textit{et al.} in the outdoor setting on the Florence dataset.

\begin{table}[t]
    \centering
    \caption[Ablation Study on AFLW2000-3D and MICC Florence Dataset]{Ablation Study on AFLW2000-3D Dataset. The “HSCA” denotes using the spatial-channel attention module or not, and the “PAF” denotes using the attention fusion module or not.}
    \label{tab:ablation}
    \begin{tabular}{lcc}
    \toprule
    \textbf{Model} & \textbf{NME on AFLW2000-3D} & \textbf{RMSE on MICC} \\
    \midrule
    our model w/o HSCA, PAF & 3.62 & $1.68 \pm 0.54$ \\
    our model w/o HSCA & 3.51 & 1.67 $\pm$ 0.45 \\
    our model w/o PAF & 3.56 & 1.67 $\pm$ 0.58 \\
    our full model & \textbf{3.41} & \textbf{1.66 $\pm$ 0.50} \\
    \bottomrule
    \end{tabular}
\end{table}

    \subsection{Ablation Study}
    To validate the effectiveness of the multi-level attention module in our proposed method MLANet, we conduct ablation experiments on the AFLW2000-3D dataset, assessing the normalized mean error (NME) under different modules. We primarily focus on the effects of the hybrid spatial-channel attention module and the progressive attention fusion module. The HSCA module introduces sequential spatial and channel self-attention within the bottleneck layers of the residual network, selectively enhancing feature representations by concentrating on key spatial regions and channel information. In contrast, the PAF module applies parallel global and local self-attention mechanisms at the network’s contextual fusion stages, achieving a balanced integration of global structural information and localized details. The ablation results, as presented in Table~\ref{tab:ablation}, illustrate the configuration of each module to the overall performance. 

    To further illustrate the impact of the multi-level attention mechanism in capturing critical facial features, we visualize feature maps generated by the HSCA module and the combined HSCA and PAF modules on selected images from the AFLW2000-3D dataset. These visualizations, based on class activation mapping (CAM) \cite{CAM}, are shown in Figure \ref{fig:grad_cam}. 
 \begin{figure}[t]
  \raggedright
  \includegraphics[width=0.9\textwidth]{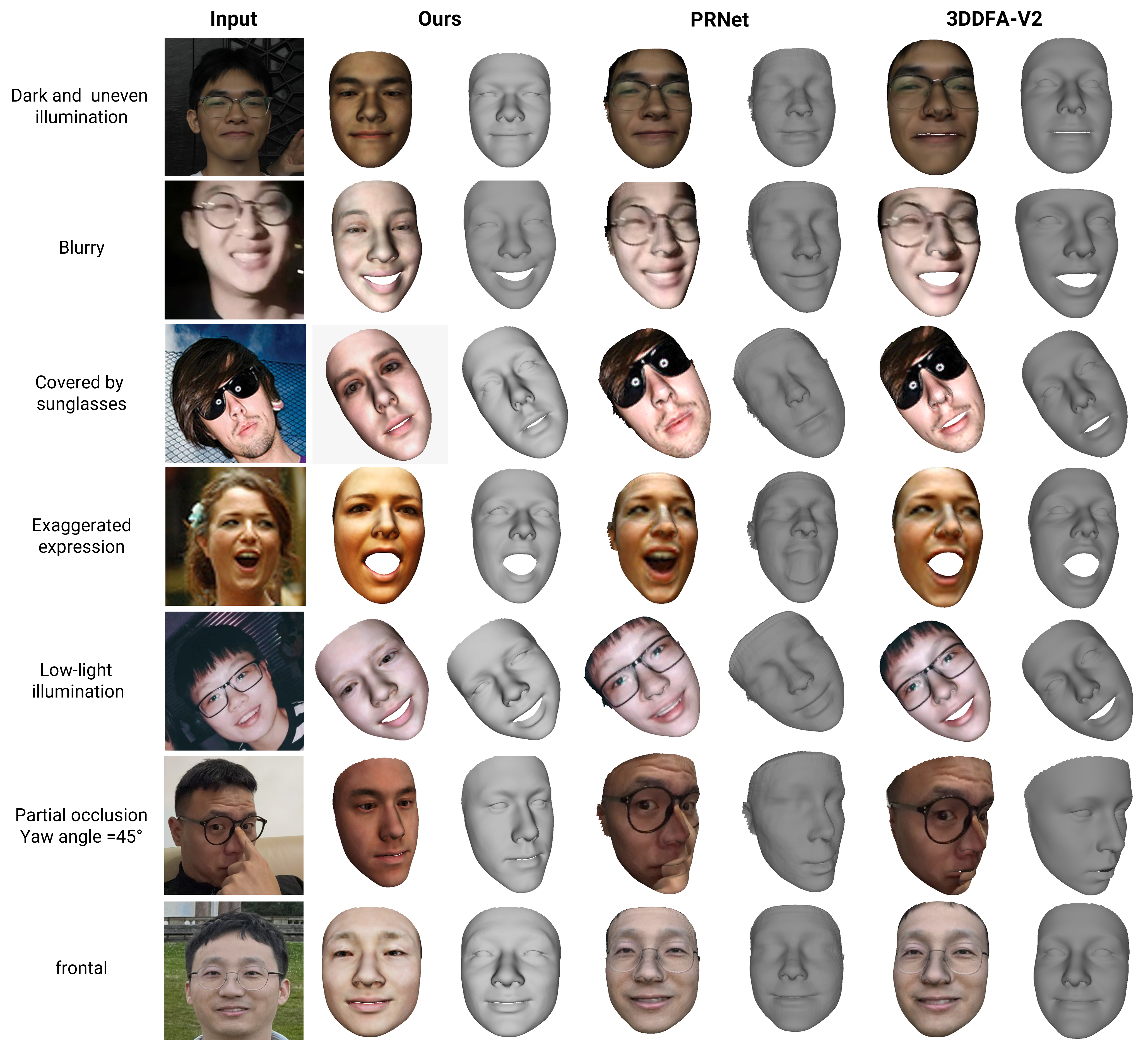}
  \caption[Qualitative comparison of 3D face reconstruction across different methods]{Visual qualitative comparison of 3D face reconstruction and rendered texture using different methods on AFLW2000-3D and a custom-collected dataset. From left to right: input image, our proposed MLANet, PRNet \cite{feng2018prn}, and 3DDFA-V2 \cite{3DDFAv2}. The rows represent various scenarios, including occlusions, different facial expressions, and large head poses.}
  \label{fig:MLANet_compar}
\end{figure} 
\subsection{Qualitative Comparison}
To qualitatively evaluate the performance of our proposed MLANet, we compared it with PRNet \cite{feng2018prn} and 3DDFA-V2 \cite{3DDFAv2} on the AFLW2000-3D dataset \cite{ALFW-3D} and a custom-collected dataset. The comparison was conducted under various scenarios, including occlusions, different facial expressions, and significant head poses, to assess the robustness and accuracy of the reconstruction methods. As shown in Figure \ref{fig:MLANet_compar}, MLANet maintained detailed and realistic 3D face reconstructions under these challenging conditions. Our method demonstrated improved resilience in images with significant occlusions, such as glasses or sunglasses, as observed in the third and sixth rows of Figure \ref{fig:MLANet_compar}. Specifically, for cases involving facial occlusions such as glasses, sunglasses, or hair partially covering the face, MLANet reconstructed the occluded regions with reasonable detail and continuity. Additionally, our approach provided more accurate geometry reconstructions in cases of pronounced facial expressions or blurred input images, as illustrated in the second and fourth rows of Figure \ref{fig:MLANet_compar}. In these scenarios, PRNet and 3DDFA-V2 tended to produce less distinct or fused facial features, such as closed or poorly defined mouth shapes. Furthermore, as shown in the first and fifth rows of Figure \ref{fig:MLANet_compar}, MLANet achieved more consistent face shapes and rendered textures under varying illumination conditions than the other baseline methods.

\begin{figure}[t]
\centering
  \includegraphics[width=0.9\textwidth]{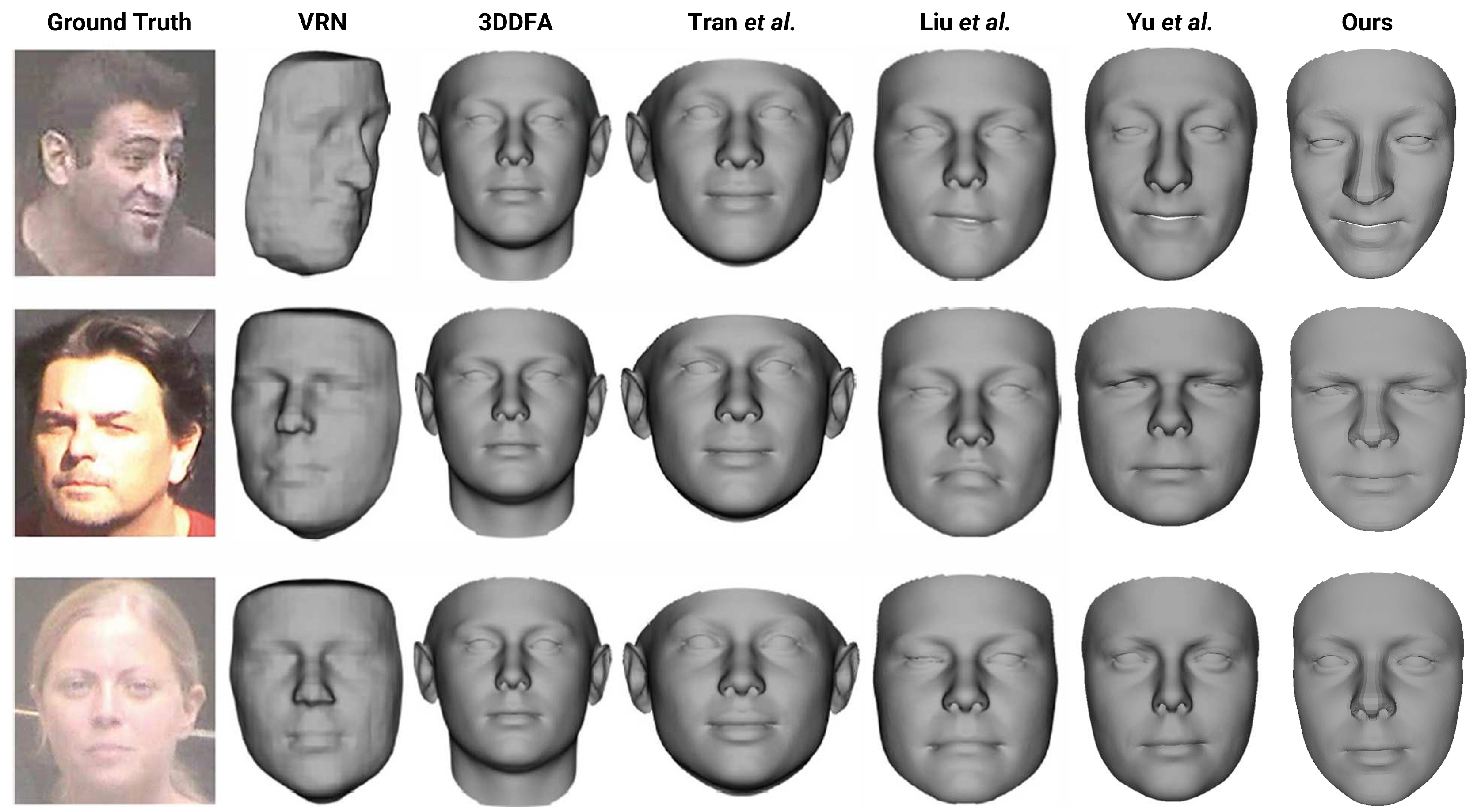}
  \caption[Qualitative comparison of face reconstruction shape on the MICC dataset]{Visual qualitative comparison of 3D face reconstruction on the MICC dataset. The comparison includes our proposed MLANet alongside VRN \cite{Jacksonvoxels3dface}, 3DDFA \cite{3DDFA}, Tran \textit{et al.} \cite{Tran_2017_CVPR}, Liu \textit{et al.} \cite{liu2018disentangling} and Yu \textit{et al.} \cite{deep3d_deng} of reconstruction shape on MICC dataset. The input images and the results of the baseline methods are sourced from Yu \textit{et al.} \cite{deep3d_deng}.}
  \label{fig:MLANet_micc}
\end{figure}

To further assess the shape reconstruction performance of MLANet, we also conducted an additional qualitative comparison on the publicly available MICC dataset, evaluating the reconstructed 3D face shapes against five baseline methods: VRN \cite{Jacksonvoxels3dface}, 3DDFA \cite{3DDFA}, Tran \textit{et al.} \cite{Tran_2017_CVPR}, Liu \textit{et al.} \cite{liu2018disentangling} and Yu \textit{et al.} \cite{deep3d_deng}. Figure \ref{fig:MLANet_micc} illustrates the learned shape for a selection of test cases. The results of the baseline methods were sourced from \cite{deep3d_deng}. It can be observed that MLANet is capable of learning more accurate shapes, including both identity and expression, particularly under challenging conditions such as large poses, exposure variations, and blurred input images.
  \section{Conclusions}
In this paper, we proposed a hierarchical multi-level attention network (MLANet) to extract deep features from face images, enhancing the accuracy and fidelity of 3D face reconstruction. MLANet integrates a hierarchical CNN backbone with multi-level attention mechanisms across different stages, progressively refining feature representations from coarse to fine. This framework strengthens the model's ability to capture complex facial structures and subtle details, even under challenging conditions.

Our training process utilizes a semi-supervised strategy, combining 3D Morphable Model parameters with a differentiable renderer to enable end-to-end learning without requiring ground-truth 3D face scans. Extensive experiments on benchmark datasets, including AFLW2000-3D and MICC Florence, demonstrate MLANet’s robustness and adaptability, achieving reliable and high-quality results in both 3D face reconstruction and alignment tasks, supported by both quantitative and qualitative evaluations.  Additionally, ablation studies validate the effectiveness of the hybrid spatial-channel attention (HSCA) and progressive attention fusion (PAF) modules, highlighting the contribution of multi-stage self-attention to improved feature representation and overall model performance.

\section*{Acknowledgments}
This work was completed during Danling's MPhil studies at the University of Manchester. Thanks to Professor Iacopo Masi and Professor Stefano Berretti, for supporting MICC Florence dataset at the University of Florence. Thanks to Danling's friends who allowed the author to collect and test their personal facial images.
\bibliographystyle{unsrt}  
\bibliography{paper}

\begin{thebibliography}{10}

\bibitem{facemodel1999}
Volker Blanz and Thomas Vetter.
\newblock A morphable model for the synthesis of 3d faces.
\newblock SIGGRAPH '99, page 187–194. ACM Press/Addison-Wesley Publishing Co., 1999.

\bibitem{MoFA_Tewari_2017_ICCV}
Ayush Tewari, Michael Zollhofer, Hyeongwoo Kim, Pablo Garrido, Florian Bernard, Patrick Perez, and Christian Theobalt.
\newblock Mofa: Model-based deep convolutional face autoencoder for unsupervised monocular reconstruction.
\newblock In {\em Proceedings of the IEEE International Conference on Computer Vision (ICCV) Workshops}, Oct 2017.

\bibitem{3DDFAv2}
Jianzhu Guo, Xiangyu Zhu, Yang Yang, Fan Yang, Zhen Lei, and Stan~Z Li.
\newblock Towards fast, accurate and stable 3d dense face alignment.
\newblock In {\em European Conference on Computer Vision}, pages 152--168. Springer, 2020.

\bibitem{deep3d_deng}
Yu~Deng, Jiaolong Yang, Sicheng Xu, Dong Chen, Yunde Jia, and Xin Tong.
\newblock Accurate 3d face reconstruction with weakly-supervised learning: From single image to image set.
\newblock In {\em Proceedings of the IEEE/CVF conference on computer vision and pattern recognition workshops}, pages 0--0, 2019.

\bibitem{MGCNet}
Jiaxiang Shang, Tianwei Shen, Shiwei Li, Lei Zhou, Mingmin Zhen, Tian Fang, and Long Quan.
\newblock Self-supervised monocular 3d face reconstruction by occlusion-aware multi-view geometry consistency.
\newblock {\em arXiv preprint arXiv:2007.12494}, 2020.

\bibitem{feng2018prn}
Yao Feng, Fan Wu, Xiaohu Shao, Yanfeng Wang, and Xi~Zhou.
\newblock Joint 3d face reconstruction and dense alignment with position map regression network.
\newblock In {\em ECCV}, 2018.

\bibitem{depth2021robust}
Peixin Li, Yuru Pei, Yicheng Zhong, Yuke Guo, and Hongbin Zha.
\newblock Robust 3d face reconstruction from single noisy depth image through semantic consistency.
\newblock {\em IET Computer Vision}, 15(6):393--404, 2021.

\bibitem{survey}
Bernhard Egger, William A.~P. Smith, Ayush Tewari, Stefanie Wuhrer, Michael Zollhoefer, Thabo Beeler, Florian Bernard, Timo Bolkart, Adam Kortylewski, Sami Romdhani, Christian Theobalt, Volker Blanz, and Thomas Vetter.
\newblock 3d morphable face models—past, present, and future.
\newblock {\em ACM Trans. Graph.}, 39(5), June 2020.

\bibitem{bfm2009}
Pascal Paysan, Reinhard Knothe, Brian Amberg, Sami Romdhani, and Thomas Vetter.
\newblock A 3d face model for pose and illumination invariant face recognition.
\newblock In {\em 2009 sixth IEEE international conference on advanced video and signal based surveillance}, pages 296--301. Ieee, 2009.

\bibitem{facewarehouse2013}
Chen Cao, Yanlin Weng, Shun Zhou, Yiying Tong, and Kun Zhou.
\newblock Facewarehouse: A 3d facial expression database for visual computing.
\newblock {\em IEEE Transactions on Visualization and Computer Graphics}, 20(3):413--425, 2013.

\bibitem{ietmaghari2014adaptive}
Ashraf Maghari, Ibrahim Venkat, Iman~Yi Liao, and Bahari Belaton.
\newblock Adaptive face modelling for reconstructing 3d face shapes from single 2d images.
\newblock {\em IET Computer Vision}, 8(5):441--454, 2014.

\bibitem{flame2017}
Tianye Li, Timo Bolkart, Michael~J Black, Hao Li, and Javier Romero.
\newblock Learning a model of facial shape and expression from 4d scans.
\newblock {\em ACM Trans. Graph.}, 36(6):194--1, 2017.

\bibitem{surrey2016}
Patrik Huber, Guosheng Hu, Rafael Tena, Pouria Mortazavian, Willem~P Koppen, William~J Christmas, Matthias R{\"a}tsch, and Josef Kittler.
\newblock A multiresolution 3d morphable face model and fitting framework.
\newblock In {\em International conference on computer vision theory and applications}, volume~5, pages 79--86. SciTePress, 2016.

\bibitem{lsfm2018}
James Booth, Anastasios Roussos, Allan Ponniah, David Dunaway, and Stefanos Zafeiriou.
\newblock Large scale 3d morphable models.
\newblock {\em International Journal of Computer Vision}, 126(2):233--254, 2018.

\bibitem{Roth_2016_CVPR}
Joseph Roth, Yiying Tong, and Xiaoming Liu.
\newblock Adaptive 3d face reconstruction from unconstrained photo collections.
\newblock In {\em Proceedings of the IEEE Conference on Computer Vision and Pattern Recognition (CVPR)}, June 2016.

\bibitem{yamaguchi2018high}
Shugo Yamaguchi, Shunsuke Saito, Koki Nagano, Yajie Zhao, Weikai Chen, Kyle Olszewski, Shigeo Morishima, and Hao Li.
\newblock High-fidelity facial reflectance and geometry inference from an unconstrained image.
\newblock {\em ACM Transactions on Graphics (TOG)}, 37(4):1--14, 2018.

\bibitem{Gecer_2019_CVPR}
Baris Gecer, Stylianos Ploumpis, Irene Kotsia, and Stefanos Zafeiriou.
\newblock Ganfit: Generative adversarial network fitting for high fidelity 3d face reconstruction.
\newblock In {\em Proceedings of the IEEE/CVF Conference on Computer Vision and Pattern Recognition (CVPR)}, June 2019.

\bibitem{Yang_2020_CVPR}
Haotian Yang, Hao Zhu, Yanru Wang, Mingkai Huang, Qiu Shen, Ruigang Yang, and Xun Cao.
\newblock Facescape: A large-scale high quality 3d face dataset and detailed riggable 3d face prediction.
\newblock In {\em Proceedings of the IEEE/CVF Conference on Computer Vision and Pattern Recognition (CVPR)}, June 2020.

\bibitem{9178990}
Yajing Chen, Fanzi Wu, Zeyu Wang, Yibing Song, Yonggen Ling, and Linchao Bao.
\newblock Self-supervised learning of detailed 3d face reconstruction.
\newblock {\em IEEE Transactions on Image Processing}, 29:8696--8705, 2020.

\bibitem{tfmeshrender_2018_CVPR}
Kyle Genova, Forrester Cole, Aaron Maschinot, Aaron Sarna, Daniel Vlasic, and William~T. Freeman.
\newblock Unsupervised training for 3d morphable model regression.
\newblock In {\em Proceedings of the IEEE Conference on Computer Vision and Pattern Recognition (CVPR)}, June 2018.

\bibitem{softrasterizer2019}
Shichen Liu, Tianye Li, Weikai Chen, and Hao Li.
\newblock Soft rasterizer: A differentiable renderer for image-based 3d reasoning.
\newblock {\em The IEEE International Conference on Computer Vision (ICCV)}, Oct 2019.

\bibitem{KaolinLibrary2022}
Clement Fuji~Tsang, Maria Shugrina, Jean~Francois Lafleche, Towaki Takikawa, Jiehan Wang, Charles Loop, Wenzheng Chen, Krishna~Murthy Jatavallabhula, Edward Smith, Artem Rozantsev, Or~Perel, Tianchang Shen, Jun Gao, Sanja Fidler, Gavriel State, Jason Gorski, Tommy Xiang, Jianing Li, Michael Li, and Rev Lebaredian.
\newblock Kaolin: A pytorch library for accelerating 3d deep learning research.
\newblock \url{https://github.com/NVIDIAGameWorks/kaolin}, 2022.

\bibitem{torch3d2020}
Justin Johnson, Nikhila Ravi, Jeremy Reizenstein, David Novotny, Shubham Tulsiani, Christoph Lassner, and Steve Branson.
\newblock Accelerating 3d deep learning with pytorch3d.
\newblock Association for Computing Machinery, 2020.

\bibitem{nvdiffrast2020}
Samuli Laine, Janne Hellsten, Tero Karras, Yeongho Seol, Jaakko Lehtinen, and Timo Aila.
\newblock Modular primitives for high-performance differentiable rendering.
\newblock {\em ACM Trans. Graph.}, 39(6), nov 2020.

\bibitem{3DDFA-v3}
Zidu Wang, Xiangyu Zhu, Tianshuo Zhang, Baiqin Wang, and Zhen Lei.
\newblock 3d face reconstruction with the geometric guidance of facial part segmentation.
\newblock In {\em Proceedings of the IEEE/CVF Conference on Computer Vision and Pattern Recognition}, pages 1672--1682, 2024.

\bibitem{withoutsupervision2019}
Soubhik Sanyal, Timo Bolkart, Haiwen Feng, and Michael Black.
\newblock Learning to regress 3d face shape and expression from an image without 3d supervision.
\newblock In {\em Proceedings IEEE Conf. on Computer Vision and Pattern Recognition (CVPR)}, June 2019.

\bibitem{EMOCA2022}
Radek Danecek, Michael~J. Black, and Timo Bolkart.
\newblock {EMOCA}: {E}motion driven monocular face capture and animation.
\newblock In {\em Conference on Computer Vision and Pattern Recognition (CVPR)}, pages 20311--20322, 2022.

\bibitem{Li_2023_CVPR}
Chunlu Li, Andreas Morel-Forster, Thomas Vetter, Bernhard Egger, and Adam Kortylewski.
\newblock Robust model-based face reconstruction through weakly-supervised outlier segmentation.
\newblock In {\em Proceedings of the IEEE/CVF Conference on Computer Vision and Pattern Recognition (CVPR)}, pages 372--381, June 2023.

\bibitem{Lin_2020_CVPR}
Jiangke Lin, Yi~Yuan, Tianjia Shao, and Kun Zhou.
\newblock Towards high-fidelity 3d face reconstruction from in-the-wild images using graph convolutional networks.
\newblock In {\em Proceedings of the IEEE/CVF Conference on Computer Vision and Pattern Recognition (CVPR)}, June 2020.

\bibitem{Gao_2020_CVPR_Workshops}
Zhongpai Gao, Juyong Zhang, Yudong Guo, Chao Ma, Guangtao Zhai, and Xiaokang Yang.
\newblock Semi-supervised 3d face representation learning from unconstrained photo collections.
\newblock In {\em Proceedings of the IEEE/CVF Conference on Computer Vision and Pattern Recognition (CVPR) Workshops}, June 2020.

\bibitem{Lee_2020_CVPR}
Gun-Hee Lee and Seong-Whan Lee.
\newblock Uncertainty-aware mesh decoder for high fidelity 3d face reconstruction.
\newblock In {\em Proceedings of the IEEE/CVF Conference on Computer Vision and Pattern Recognition (CVPR)}, June 2020.

\bibitem{Qiu_2021_CVPR}
Yuda Qiu, Xiaojie Xu, Lingteng Qiu, Yan Pan, Yushuang Wu, Weikai Chen, and Xiaoguang Han.
\newblock 3dcaricshop: A dataset and a baseline method for single-view 3d caricature face reconstruction.
\newblock In {\em Proceedings of the IEEE/CVF Conference on Computer Vision and Pattern Recognition (CVPR)}, pages 10236--10245, June 2021.

\bibitem{Tran_2017_CVPR}
Anh Tuan~Tran, Tal Hassner, Iacopo Masi, and Gerard Medioni.
\newblock Regressing robust and discriminative 3d morphable models with a very deep neural network.
\newblock In {\em Proceedings of the IEEE Conference on Computer Vision and Pattern Recognition (CVPR)}, July 2017.

\bibitem{Yi_2019_CVPR}
Hongwei Yi, Chen Li, Qiong Cao, Xiaoyong Shen, Sheng Li, Guoping Wang, and Yu-Wing Tai.
\newblock Mmface: A multi-metric regression network for unconstrained face reconstruction.
\newblock In {\em Proceedings of the IEEE/CVF Conference on Computer Vision and Pattern Recognition (CVPR)}, June 2019.

\bibitem{MICA2022}
{\em Towards Metrical Reconstruction of Human Faces}, 2022.

\bibitem{resnet2016}
Kaiming He, Xiangyu Zhang, Shaoqing Ren, and Jian Sun.
\newblock Deep residual learning for image recognition.
\newblock In {\em Proceedings of the IEEE conference on computer vision and pattern recognition}, pages 770--778, 2016.

\bibitem{fpn2017}
Tsung-Yi Lin, Piotr Doll{\'a}r, Ross Girshick, Kaiming He, Bharath Hariharan, and Serge Belongie.
\newblock Feature pyramid networks for object detection.
\newblock In {\em Proceedings of the IEEE conference on computer vision and pattern recognition}, pages 2117--2125, 2017.

\bibitem{ViTsurvey_hankai}
Kai Han, Yunhe Wang, Hanting Chen, Xinghao Chen, Jianyuan Guo, Zhenhua Liu, Yehui Tang, An~Xiao, Chunjing Xu, Yixing Xu, et~al.
\newblock A survey on vision transformer.
\newblock {\em IEEE transactions on pattern analysis and machine intelligence}, 45(1):87--110, 2022.

\bibitem{Imagesr2023}
Image super-resolution: A comprehensive review, recent trends, challenges and applications.
\newblock {\em Information Fusion}, 91:230--260, 2023.

\bibitem{imgrecog2020}
Hengshuang Zhao, Jiaya Jia, and Vladlen Koltun.
\newblock Exploring self-attention for image recognition.
\newblock In {\em Proceedings of the IEEE/CVF conference on computer vision and pattern recognition}, pages 10076--10085, 2020.

\bibitem{attimgcls2023}
Haimiao Ge, Liguo Wang, Moqi Liu, Xiaoyu Zhao, Yuexia Zhu, Haizhu Pan, and Yanzhong Liu.
\newblock Pyramidal multiscale convolutional network with polarized self-attention for pixel-wise hyperspectral image classification.
\newblock {\em IEEE Transactions on Geoscience and Remote Sensing}, 61:1--18, 2023.

\bibitem{segattention2022}
Shumin An, Qingmin Liao, Zongqing Lu, and Jing-Hao Xue.
\newblock Efficient semantic segmentation via self-attention and self-distillation.
\newblock {\em IEEE Transactions on Intelligent Transportation Systems}, 23(9):15256--15266, 2022.

\bibitem{FsaNet2023}
Fengyu Zhang, Ashkan Panahi, and Guangjun Gao.
\newblock Fsanet: Frequency self-attention for semantic segmentation.
\newblock {\em IEEE Transactions on Image Processing}, 32:4757--4772, 2023.

\bibitem{senet2018}
Jie Hu, Li~Shen, and Gang Sun.
\newblock Squeeze-and-excitation networks.
\newblock In {\em Proceedings of the IEEE conference on computer vision and pattern recognition}, pages 7132--7141, 2018.

\bibitem{cbam2018}
Sanghyun Woo, Jongchan Park, Joon-Young Lee, and In~So Kweon.
\newblock Cbam: Convolutional block attention module.
\newblock In {\em Proceedings of the European conference on computer vision (ECCV)}, pages 3--19, 2018.

\bibitem{hu2018genet}
Jie Hu, Li~Shen, Samuel Albanie, Gang Sun, and Andrea Vedaldi.
\newblock Gather-excite: Exploiting feature context in convolutional neural networks.
\newblock 2018.

\bibitem{misra2021TA}
Diganta Misra, Trikay Nalamada, Ajay~Uppili Arasanipalai, and Qibin Hou.
\newblock Rotate to attend: Convolutional triplet attention module.
\newblock In {\em Proceedings of the IEEE/CVF winter conference on applications of computer vision}, pages 3139--3148, 2021.

\bibitem{SH2001}
Ravi Ramamoorthi and Pat Hanrahan.
\newblock A signal-processing framework for inverse rendering.
\newblock In {\em Proceedings of the 28th annual conference on Computer graphics and interactive techniques}, pages 117--128, 2001.

\bibitem{AFF2021}
Yimian Dai, Fabian Gieseke, Stefan Oehmcke, Yiquan Wu, and Kobus Barnard.
\newblock Attentional feature fusion.
\newblock In {\em Proceedings of the IEEE/CVF winter conference on applications of computer vision}, pages 3560--3569, 2021.

\bibitem{saito2016real}
Shunsuke Saito, Tianye Li, and Hao Li.
\newblock Real-time facial segmentation and performance capture from rgb input.
\newblock In {\em Computer Vision--ECCV 2016: 14th European Conference, Amsterdam, The Netherlands, October 11-14, 2016, Proceedings, Part VIII 14}, pages 244--261. Springer, 2016.

\bibitem{arcface}
Jiankang Deng, Jia Guo, Niannan Xue, and Stefanos Zafeiriou.
\newblock Arcface: Additive angular margin loss for deep face recognition.
\newblock In {\em Proceedings of the IEEE/CVF Conference on Computer Vision and Pattern Recognition}, pages 4690--4699, 2019.

\bibitem{ietland2019survey}
El~Rhazi Manal, Zarghili Arsalane, and Majda Aicha.
\newblock Survey on the approaches based geometric information for 3d face landmarks detection.
\newblock {\em IET Image Processing}, 13(8):1225--1231, 2019.

\bibitem{bulat2017far}
Adrian Bulat and Georgios Tzimiropoulos.
\newblock How far are we from solving the 2d \& 3d face alignment problem? (and a dataset of 230,000 3d facial landmarks).
\newblock In {\em International Conference on Computer Vision}, 2017.

\bibitem{tewari2018self}
Ayush Tewari, Michael Zollh{\"o}fer, Pablo Garrido, Florian Bernard, Hyeongwoo Kim, Patrick P{\'e}rez, and Christian Theobalt.
\newblock Self-supervised multi-level face model learning for monocular reconstruction at over 250 hz.
\newblock In {\em Proceedings of the IEEE conference on computer vision and pattern recognition}, pages 2549--2559, 2018.

\bibitem{mtcnn}
Kaipeng Zhang, Zhanpeng Zhang, Zhifeng Li, and Yu~Qiao.
\newblock Joint face detection and alignment using multitask cascaded convolutional networks.
\newblock {\em IEEE signal processing letters}, 23(10):1499--1503, 2016.

\bibitem{celebA}
Ziwei Liu, Ping Luo, Xiaogang Wang, and Xiaoou Tang.
\newblock Deep learning face attributes in the wild.
\newblock In {\em Proceedings of International Conference on Computer Vision (ICCV)}, December 2015.

\bibitem{3DDFA}
Xiangyu Zhu, Xiaoming Liu, Zhen Lei, and Stan~Z. Li.
\newblock Face alignment in full pose range: A 3d total solution.
\newblock {\em IEEE Transactions on Pattern Analysis and Machine Intelligence}, 41(1):78–92, January 2019.

\bibitem{skindetection}
Michael~J Jones and James~M Rehg.
\newblock Statistical color models with application to skin detection.
\newblock {\em International journal of computer vision}, 46:81--96, 2002.

\bibitem{ALFW-3D}
Xiangyu Zhu, Zhen Lei, Xiaoming Liu, Hailin Shi, and Stan~Z Li.
\newblock Face alignment across large poses: A 3d solution.
\newblock In {\em Proceedings of the IEEE conference on computer vision and pattern recognition}, pages 146--155, 2016.

\bibitem{Florence}
Andrew~D. Bagdanov, Alberto Del~Bimbo, and Iacopo Masi.
\newblock The florence 2d/3d hybrid face dataset.
\newblock In {\em Proceedings of the 2011 Joint ACM Workshop on Human Gesture and Behavior Understanding}, page 79–80. ACM, 2011.

\bibitem{resnet2015imagenet}
Olga Russakovsky, Jia Deng, Hao Su, Jonathan Krause, Sanjeev Satheesh, Sean Ma, Zhiheng Huang, Andrej Karpathy, Aditya Khosla, Michael Bernstein, et~al.
\newblock Imagenet large scale visual recognition challenge.
\newblock {\em International journal of computer vision}, 115:211--252, 2015.

\bibitem{adam}
DP~Kingma, LJ~Ba, et~al.
\newblock Adam: A method for stochastic optimization.
\newblock 2015.

\bibitem{SADRNET}
Zeyu Ruan, Changqing Zou, Longhai Wu, Gangshan Wu, and Limin Wang.
\newblock Sadrnet: Self-aligned dual face regression networks for robust 3d dense face alignment and reconstruction.
\newblock {\em Trans. Img. Proc.}, 30:5793–5806, January 2021.

\bibitem{SDM2015}
Xuehan Xiong and Fernando De~la Torre.
\newblock Global supervised descent method.
\newblock In {\em Proceedings of the IEEE Conference on Computer Vision and Pattern Recognition}, pages 2664--2673, 2015.

\bibitem{CPEM}
Langyuan Mo, Haokun Li, Chaoyang Zou, Yubing Zhang, Ming Yang, Yihong Yang, and Mingkui Tan.
\newblock Towards accurate facial motion retargeting with identity-consistent and expression-exclusive constraints.
\newblock In {\em Proceedings of the AAAI Conference on Artificial Intelligence}, volume~36, pages 1981--1989, 2022.

\bibitem{booth20173d}
James Booth, Epameinondas Antonakos, Stylianos Ploumpis, George Trigeorgis, Yannis Panagakis, and Stefanos Zafeiriou.
\newblock 3d face morphable models" in-the-wild".
\newblock In {\em Proceedings of the IEEE conference on computer vision and pattern recognition}, pages 48--57, 2017.

\bibitem{genova2018unsupervised}
Kyle Genova, Forrester Cole, Aaron Maschinot, Aaron Sarna, Daniel Vlasic, and William~T Freeman.
\newblock Unsupervised training for 3d morphable model regression.
\newblock In {\em Proceedings of the IEEE conference on computer vision and pattern recognition}, pages 8377--8386, 2018.

\bibitem{ICP}
Paul~J. Besl and Neil~D. McKay.
\newblock {Method for registration of 3-D shapes}.
\newblock In Paul~S. Schenker, editor, {\em Sensor Fusion IV: Control Paradigms and Data Structures}, volume 1611, pages 586 -- 606. International Society for Optics and Photonics, SPIE, 1992.

\bibitem{MM}
Qixin Deng, Binh~H. Le, Aobo Jin, and Zhigang Deng.
\newblock End-to-end 3d face reconstruction with expressions and specular albedos from single in-the-wild images.
\newblock In {\em Proceedings of the 30th ACM International Conference on Multimedia}, MM '22, page 4694–4703. Association for Computing Machinery, 2022.

\bibitem{CAM}
Ramprasaath~R Selvaraju, Michael Cogswell, Abhishek Das, Ramakrishna Vedantam, Devi Parikh, and Dhruv Batra.
\newblock Grad-cam: visual explanations from deep networks via gradient-based localization.
\newblock {\em International journal of computer vision}, 128:336--359, 2020.

\bibitem{Jacksonvoxels3dface}
Aaron~S Jackson, Adrian Bulat, Vasileios Argyriou, and Georgios Tzimiropoulos.
\newblock Large pose 3d face reconstruction from a single image via direct volumetric cnn regression.
\newblock In {\em Proceedings of the IEEE international conference on computer vision}, pages 1031--1039, 2017.

\bibitem{liu2018disentangling}
Feng Liu, Ronghang Zhu, Dan Zeng, Qijun Zhao, and Xiaoming Liu.
\newblock Disentangling features in 3d face shapes for joint face reconstruction and recognition.
\newblock In {\em Proceedings of the IEEE conference on computer vision and pattern recognition}, pages 5216--5225, 2018.

\end{thebibliography}

\end{document}